\documentclass[journal]{IEEEtran}

\usepackage{threeparttable}
\usepackage{graphicx}
\usepackage{caption}
\usepackage{subcaption}
\usepackage{amsmath}
\usepackage{amssymb}

\def\etal{\emph{et al}.}

\usepackage{diagbox}
\usepackage{algorithm}
\usepackage{algorithmic}

\usepackage[T1]{fontenc}
\usepackage{graphicx}
\usepackage{amsmath}
\usepackage{amssymb}
\usepackage{booktabs}
\usepackage{multirow}
\usepackage{bbm}
\usepackage{pifont}
\usepackage{enumitem}
\usepackage{listings}
\usepackage{amsfonts}
\usepackage{nicefrac}
\usepackage{microtype}
\usepackage{float}
\usepackage{caption}
\usepackage{wrapfig}
\usepackage{lipsum}
\usepackage[dvipsnames]{xcolor}
\usepackage[accsupp]{axessibility}
\usepackage{sidecap}
\usepackage{tabulary}

\newcolumntype{y}[1]{>{\raggedright\arraybackslash}p{#1pt}}
\newcolumntype{z}[1]{>{\raggedleft\arraybackslash}p{#1pt}}

\setlist[enumerate,1]{%
  label=\arabic*.,
}

\newlist{inlinelist}{enumerate*}{1}
\setlist*[inlinelist,1]{%
  label=(\roman*),
}

\definecolor{citecolor}{HTML}{0071bc}
\definecolor{bluegreen}{RGB}{117, 171, 188}

\usepackage[pagebackref,breaklinks,colorlinks, anchorcolor=blue, citecolor=citecolor]{hyperref}

\usepackage[capitalize]{cleveref}
\crefname{section}{Sec.}{Secs.}
\Crefname{section}{Section}{Sections}
\Crefname{table}{Table}{Tables}
\crefname{table}{Tab.}{Tabs.}

\newlength\savewidth
\newcommand{\pub}[1]{\color{gray}{\tiny{[{#1}]}}}

\begin{document}
%
% paper title
% Titles are generally capitalized except for words such as a, an, and, as,
% at, but, by, for, in, nor, of, on, or, the, to and up, which are usually
% not capitalized unless they are the first or last word of the title.
% Linebreaks \\ can be used within to get better formatting as desired.
% Do not put math or special symbols in the title.
\title{Harnessing Group-Oriented Consistency Constraints for Semi-Supervised Semantic Segmentation in CdZnTe Semiconductors}

\author{Peihao~Li,
Yan~Fang,
Man~Liu,
Huihui~Bai,
Anhong~Wang,
Yunchao~Wei,
and~Yao~Zhao,~\IEEEmembership{Fellow,~IEEE}
%\author{Michael~Shell,~\IEEEmembership{Member,~IEEE,}
%  John~Doe,~\IEEEmembership{Fellow,~OSA,}
%  and~Jane~Doe,~\IEEEmembership{Life~Fellow,~IEEE}% <-this % stops a space

\thanks{ 
Peihao~Li, Yan~Fang, Huihui Bai, Yunchao Wei, and Yao Zhao are with the Institute of Information Science, Beijing Key Laboratory of Advanced Information Science and Network Technology, Beijing Jiaotong University, Beijing 100082, China (e-mail: lphao\_g@bjtu.edu.cn; 22110080@bjtu.edu.cn; hhbai@bjtu.edu.cn; yunchao.wei@bjtu.edu.cn; yzhao@bjtu.edu.cn). (Corresponding authors: Huihui Bai; Man Liu.)

Man Liu is with the School of Artificial Intelligence, Anhui University, Hefei, 230601, China (e-mail: manliu@ahu.edu.cn).

Anhong Wang is with the Institute of Digital Media Communication, Taiyuan University of Science and Technology, Taiyuan 030024, China (e-mail: ahwang@tyust.edu.cn).}
}

\maketitle
\begin{abstract}

Labeling Cadmium Zinc Telluride (CdZnTe) semiconductor images is challenging due to the low-contrast defect boundaries, necessitating annotators to cross-reference multiple views. These views share a single ground truth (GT), forming a unique ``many-to-one'' relationship. This characteristic renders advanced semi-supervised semantic segmentation (SSS) methods suboptimal, as they are generally limited by a ``one-to-one'' relationship, where each image is independently associated with its GT. Such limitation may lead to error accumulation in low-contrast regions, further exacerbating confirmation bias. To address this issue, we revisit the SSS pipeline from a group-oriented perspective and propose a human-inspired solution: the Intra-group Consistency Augmentation Framework (ICAF). First, we experimentally validate the inherent consistency constraints within CdZnTe groups, establishing a group-oriented baseline using the Intra-group View Sampling (IVS). Building on this insight, we introduce the Pseudo-label Correction Network (PCN) to enhance consistency representation, which consists of two key modules. The View Augmentation Module (VAM) improves boundary details by dynamically synthesizing a boundary-aware view through the aggregation of multiple views. In the View Correction Module (VCM), this synthesized view is paired with other views for information interaction, effectively emphasizing salient regions while minimizing noise. Extensive experiments demonstrate the effectiveness of our solution for CdZnTe materials. Leveraging DeepLabV3+ with a ResNet-101 backbone as our segmentation model, we achieve a 70.6\% mIoU on the CdZnTe dataset using only 2 group-annotated data (5\textperthousand). The code is available at \href{https://github.com/pipixiapipi/ICAF}{https://github.com/pipixiapipi/ICAF}.

\end{abstract}


\IEEEpeerreviewmaketitle

\section{Introduction}  

\IEEEPARstart{I}{n} recent years, advancements in computational power, model optimization, and access to expansive datasets have markedly accelerated progress in deep learning~\cite{SSS_UPC}. These developments, particularly in end-to-end training methodologies, have been extensively adopted across a wide array of industrial applications~\cite{Wavelet, lightweight, GAN_1}. Nevertheless, the inherent complexities of industrial production environments, coupled with the diverse range of defect types, render the low-contrast challenge one of the most critical obstacles in automated segmentation~\cite{challenge}.

\begin{figure}[t]
\centering
  \includegraphics[width=\linewidth]{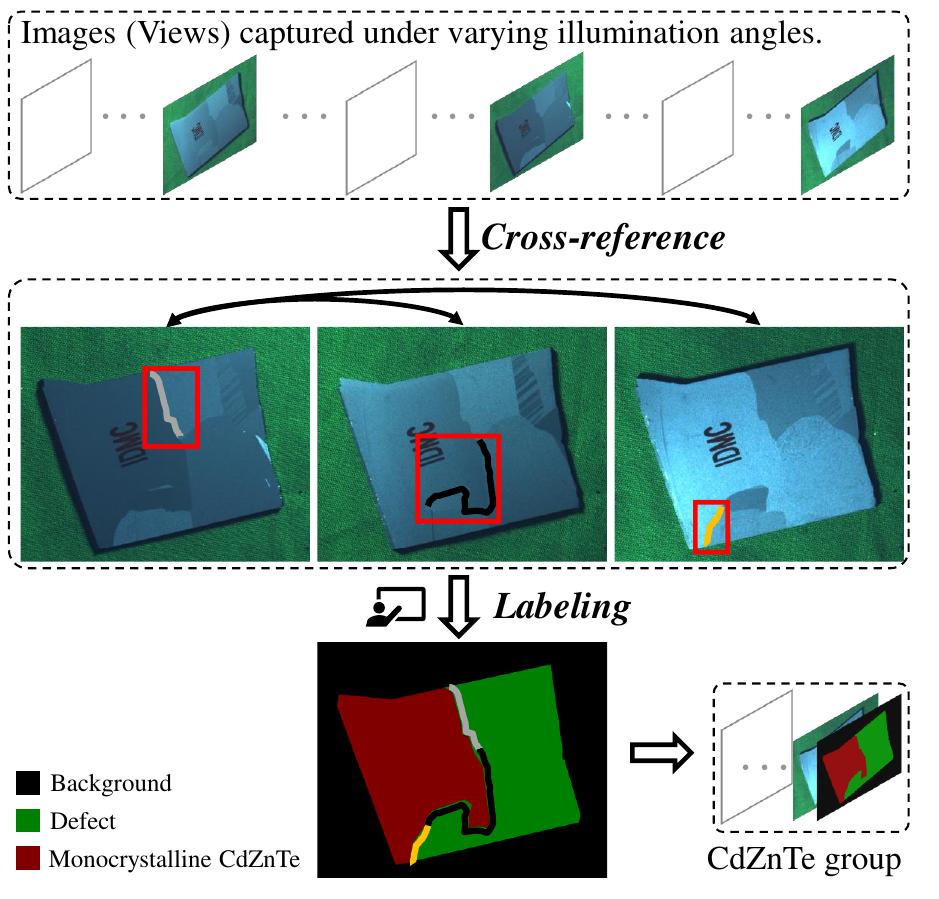}
\caption{The labeling pipeline for a CdZnTe sample. The {\color{red} red} rectangular boxes delineate the cross-referenced regions.
\label{image_TPO}}
\end{figure}
\setlength{\textfloatsep}{5pt} 

Cadmium Zinc Telluride (CdZnTe) semiconductors are advanced materials used in infrared, X-ray, and high-energy radiation detectors. Fig. \ref{image_TPO} demonstrates significant variations in defect boundary appearance when imaging the same CdZnTe sample under different illumination angles. These inconsistencies arise from the intricate surface textures and varying reflective properties of CdZnTe. Such material-specific characteristics significantly exacerbate the low-contrast detection challenges in CdZnTe semiconductors. To achieve precise defect boundary delineation, annotators need to cross-reference multiple images (views) captured under various lighting angles, which are linked to a single ground truth (GT) label, establishing a ``many-to-one'' correspondence. This multi-view relationship constitutes the CdZnTe group for each sample. However, this cross-referencing process for high-quality labeling within high-resolution CdZnTe groups is both labor-intensive and costly.

To reduce reliance on exhaustive annotations, weakly-supervised~\cite{weak}, semi-supervised~\cite{SSS_Unimatch}, and unsupervised~\cite{unsupervised} semantic segmentation methods have emerged as promising alternatives. Given the large-scale and continuous nature of manufacturing processes, vast amounts of unlabeled data can be easily acquired. This makes semi-supervised semantic segmentation (SSS) an efficient strategy to bridge the gap between the scarcity of finely labeled data and the abundance of readily accessible unlabeled data.

\begin{figure}
\centering\includegraphics[width=\linewidth]{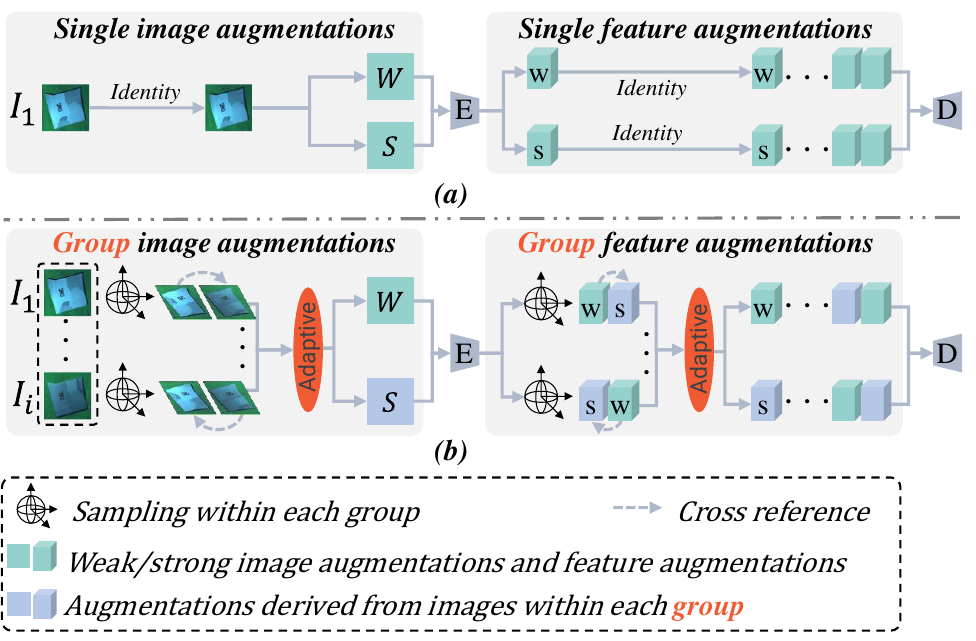}
\caption{Comparison of the traditional training pipeline and our group-oriented solution. (a) The traditional pipeline relies solely on various augmentations derived from the data itself. (b) The proposed group-oriented architecture cross-references multiple views within each group to facilitate adaptive view interaction.}
\label{comparison}
\end{figure}

Most advanced SSS methods~\cite{SSS_Unimatch, SSS_Corrmatch} typically assume a ``one-to-one'' correspondence, as seen in datasets like Pascal~\cite{dataset_pascal}, where each image is mapped to a single GT. As shown in Fig. \ref{comparison}{\color{red}(a)}, these methods rely solely on image-wise and feature-wise augmentations derived from the data itself, rendering them suboptimal for low-contrast industrial environments, as they lack the capability to manage the ``many-to-one'' correspondence within groups. This unique correspondence is a double-edged sword. On one hand, a group of views provides rich boundary information that potentially enhances segmentation accuracy. On the other hand, the significant variation among views sharing the same GT may confuse the segmentation model, leading to error accumulation and overfitting to incorrect pseudo-labels predicted by the network, a phenomenon known as confirmation bias~\cite{confirmation_bias}. This presents a challenging question:

\textbf{How can we effectively harness the abundant information embedded in different views while minimizing error accumulation and mitigating confirmation bias?}

As illustrated in Fig. \ref{image_TPO}, human annotators often cross-reference multiple views to delineate complete defect boundaries, especially in low-contrast regions—this practice inspires our approach. To emulate this strategy, we revisit the SSS pipeline from a group-oriented perspective, incorporating adaptive image and feature augmentations to facilitate information interaction within each group, as shown in Fig.~\ref{comparison}{\color{red}(b)}. Specifically, we propose the Intra-group View Sampling (IVS), which establishes a new group-baseline by exploring the inherent consistency constraints within the CdZnTe group. Building on this foundation, we introduce the Pseudo-label Correction Network (PCN) to further enhance consistency within the group by improving the quality of pseudo-labels. The PCN consists of two key modules: the View Augmentation Module (VAM) and the View Correction Module (VCM). VAM utilizes a weight generation unit to aggregate boundary clues from multiple views, generating a boundary-aware view to refine boundary details. Then, this boundary-aware view is paired with other views within the same group and fed into the VCM, which employs a spatial interaction unit to exchange view information, focusing on key regions while minimizing noise. These group-oriented strategies, specifically designed for the ``many-to-one'' relationship of CdZnTe semiconductors, collectively constitute the Intra-group Consistency Augmentation Framework (ICAF). 

In summary, the main contributions of this paper are threefold: 

\begin{itemize}

\item To the best of our knowledge, we are the first to address the SSS from a group-oriented perspective, and we have demonstrated the effectiveness of our solution on CdZnTe materials.

\item We propose a human-inspired solution, ICAF, which consists of IVS and PCN to explore correlations within the CdZnTe group. IVS is employed to establish consistency constraints, and PCN further enhances consistency representation by generating the boundary-aware view via VAM and implementing boundary correction through view interaction via VCM.

\item Extensive experiments demonstrate that our proposed ICAF significantly mitigates the effects of confirmation bias, achieving state-of-the-art performance in automated defect segmentation for CdZnTe materials. Notably, this improvement is achieved without adding any additional computational burden during inference.

\end{itemize}

\begin{figure*}[t]
\centering\includegraphics[scale=0.6]{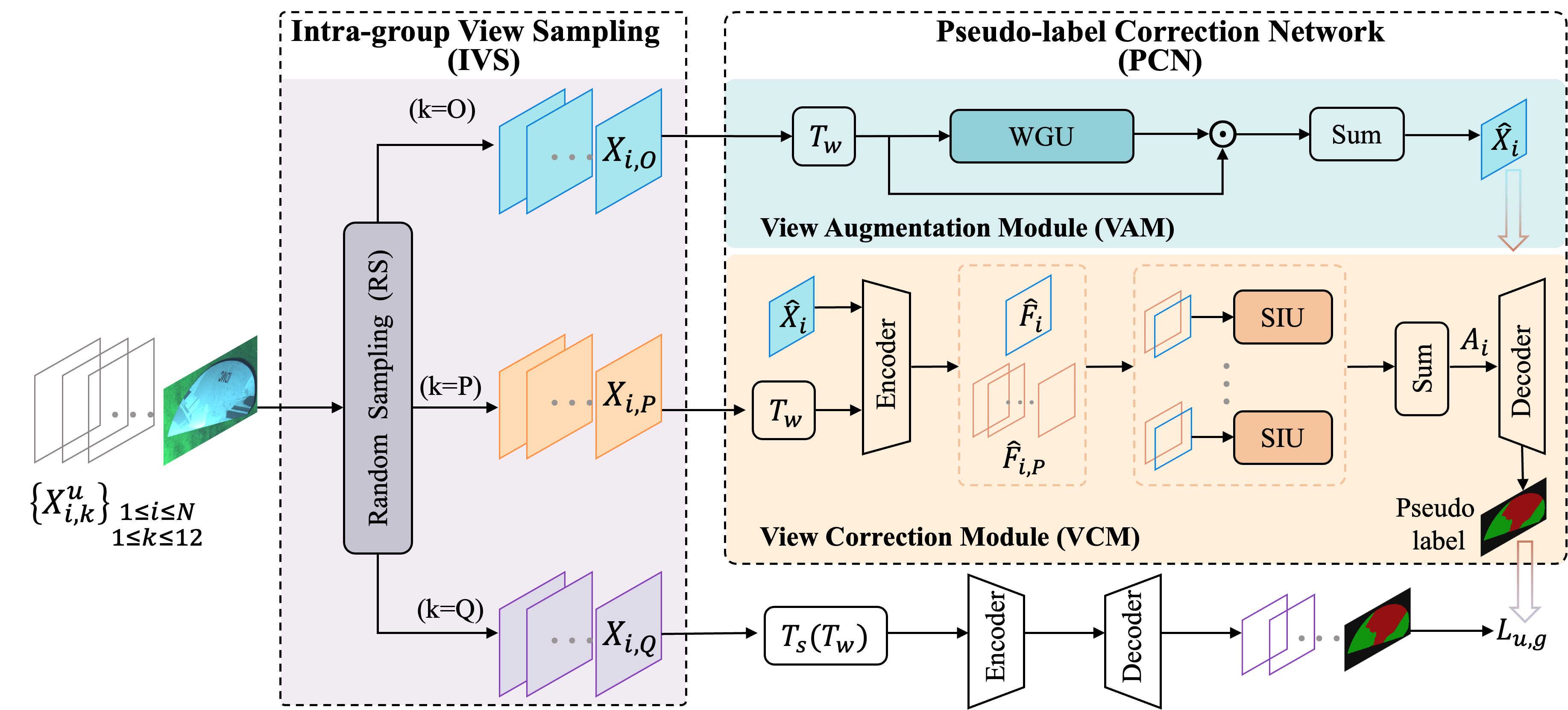}
\caption{The pipeline of our Intra-group Consistency Augmentation Framework (ICAF) for unlabeled data begins with the implementation of Intra-group View Sampling (IVS) to establish a weak-to-strong consistency constraint from a group-oriented perspective. Building upon this foundation, we employ the Pseudo-label Correction Network (PCN) to mitigate confirmation bias in pseudo-labels. Specifically, the View Augmentation Module (VAM) enhances boundary details by aggregating information from multiple views to generate the boundary-aware view $\hat{X}_{i}$, and the View Correction Module (VCM) further eliminates representation redundancy by processing each pair of $\hat{X}_{i}$ and $X_{i,P}$. The $O$, $P$ and $Q$ represent the number of views sampled from the CdZnTe group for different modules, where $P, Q \le O$. $T_w$ and $T_s$ denote weak geometric augmentation and strong intensity-based augmentation, respectively. For labeled data, we perform the same operations, excluding the $Q$ strongly perturbed branches. }
\label{framework}
\end{figure*}
 
\section{Related Work}

This section presents an overview of defect segmentation techniques used in industrial applications, introduces several semi-supervised learning frameworks, and reviews recent advancements in SSS.

\subsection{Defect Segmentation Methods}

In the past, quality inspection tasks relied heavily on manual visual checks, which are increasingly insufficient to meet the high standards of modern manufacturing. Traditional computer vision algorithms~\cite{Traditional1, Traditional2} have often struggled with the complex texture patterns and low-contrast conditions typical of industrial environments, requiring extensive task-specific feature engineering. These methods result in complicated and inefficient workflows. However, the advent of deep learning, particularly convolutional neural networks (CNNs), has revolutionized defect segmentation, significantly improving both accuracy and efficiency in industrial applications. Recent research on CNN-based defect inspection can be categorized into different primary directions: multi-source information fusion~\cite{lightweight, PCKA, GMFM}, defect generation for data augmentation~\cite{GAN_1, GAN_2, GAN_3}, and advanced architectural innovations leveraging multi-task learning~\cite{Multi_Task}. For example, Du \etal~\cite{GMFM} utilized a Gated Multilayer Fusion Module (GMFM) to merge original and CLAHE-processed X-ray images, thus enhancing segmentation accuracy for automotive casting aluminum parts. Yang \etal~\cite{GAN_3} presented a Semantic Information Decomposition Network (SIDN) to effectively segment various texture defects. Sampath \etal~\cite{Multi_Task} proposed Defect-Aux-Net, an attention-guided multitask learning framework that improves defect classification, detection, and segmentation performance. Recently, attention-based architectures, particularly Vision Transformers~\cite{VIT}, have emerged as powerful alternatives to CNNs due to their global receptive field and context modeling capabilities. In this paradigm, Zhang \etal~\cite{Wavelet} introduced a two-stage Promotion-Suppression Transformer (PST) framework that leverages wavelet features to enhance the network's capacity for capturing detailed image characteristics. 

Despite these advances, existing methods still heavily rely on precise expert annotations, which are costly and time-consuming to produce. This challenge has driven us to explore utilizing large volumes of unlabeled data alongside a limited set of labeled data to train models more effectively.

\subsection{Semi-Supervised Learning}

Semi-supervised learning (SSL) has gained significant attention due to its capability to leverage large amounts of unlabeled data, thus reducing the dependence on labeled examples. Research in SSL has primarily focused on two main approaches: self-training~\cite{ST} and consistency regularization~\cite{CR_MT, mixmatch, CR_UDA, CR_fixmatch, CR_flexmatch}. Self-training methodology entails the initial training of a model using labeled data, followed by the generation of pseudo-labels for unlabeled instances, which are subsequently utilized as GT annotations for further model optimization. However, the complex pipeline associated with self-training limits its practical applicability. The other solution is consistency regularization, which seeks to improve model robustness by ensuring consistent predictions across different input perturbations. The other solution is consistency regularization, which aims to enhance model robustness by enforcing consistent predictions across diverse input perturbations. FixMatch~\cite{CR_fixmatch} introduced a novel consistency learning method that uses weakly augmented images to guide the training for their strongly augmented counterparts, consequently yielding substantial improvements in SSL performance. FlexMatch~\cite{CR_flexmatch} extended this consistency paradigm by employing class-specific thresholds to filter out low-confidence predictions, thereby refining the pseudo-labeling process. 

\subsection{Semi-Supervised Semantic Segmentation}

Building upon the advancements in SSL, SSS aims to reduce the burden of acquiring pixel-level annotations~\cite{SSS_UPC, SSS_Unimatch, SSS_Corrmatch, SSS_Augmentation, SSS_ELN, SSS_St++, SSS_conservative, SSS_imas, SSS_U2PL, SSS_Adversarial, SSS_perspective}. This approach is particularly crucial in domains such as industrial manufacturing \cite{SSS_Semicurv} and medical imaging \cite{SSS_gland}, where the acquisition of labeled data is both challenging and costly. Early approaches to SSS often employed generative adversarial networks (GANs) to generate supervisory signals by deceiving the discriminator~\cite{SSS_GAN}. However, the inherent instability of GAN training led researchers to explore more reliable alternatives, such as consistency regularization~\cite{SSS_Unimatch, SSS_Corrmatch, SSS_cct, SSS_Pseudoseg} and self-training~\cite{SSS_St++, SSS_three, SSS_baseline}. These two methods are frequently combined in the SSS field to tackle the inherently complex task of pixel-level classification. For example, UPC~\cite{SSS_UPC} introduced an uncertainty map-based denoising strategy to refine pseudo-labels and improve segmentation performance. Inspired by FixMatch, UniMatch~\cite{SSS_Unimatch} adopted a weak-to-strong consistency training approach, establishing a new benchmark for SSS. Based on UniMatch, CorrMatch~\cite{SSS_Corrmatch} advanced the field by incorporating label propagation via correlation matching, effectively harnessing the semantic relationships present in unlabeled data. Beyond these paradigms, Transformer architectures have been explored within the SSS domain. Allspark \cite{SSS_spark} introduced a fully Transformer-based framework that effectively mitigates the limitations stemming from excessive dependence on labeled training data.

Despite advancements in state-of-the-art SSS methods, most approaches remain restricted by a ``one-to-one'' mapping between input data and GT. This limitation hinders their ability to handle the ``many-to-one'' structure inherent in the CdZnTe semiconductor dataset, as they fail to capture correlations among multiple images within the same group, resulting in suboptimal performance. To address this, we propose a human-inspired approach that integrates multi-input relationships into the SSS pipeline, effectively mimicking human cognitive processes to cross-reference views and delineate complete defect boundaries.

\section{Approach}

\begin{algorithm}[t]
\caption{Pseudocode of ICAF in a PyTorch-like style.}
\label{algorithm}

\begin{algorithmic}[1] %[1] enables line numbers

\FOR{$(X_{i,O}, X_{i,P}, X_{i,Q})$ \textbf{in} loader\_u}

\STATE \COMMENT{View enhancement (O=6)}
\STATE $\hat{X}_{i} = VAM(T_w(X_{i,O}))$ 

\STATE \COMMENT{View interaction (P=3)}
\STATE $\hat{F}_{i}, F_{i,P} = Encoder(\hat{X}_{i}, T_w(X_{i,P}))$ 
\STATE $A_i = \sum_{p=1}^{P} VCM(\hat{F}_{i}, F_{i,p})$

\STATE \COMMENT{Pseudo-label generation}
\STATE $ p_{pseudo} = Decoder(A_i).argmax(dim=1)$

\STATE \COMMENT{Consistency constraint (Q=2)}
\STATE $ p_{fa} = Decoder(nn.Dropout2d(A_i))$
\STATE $p_{i,1}, p_{i,2} = Decoder(Encoder(T_s(T_w(X_{i,Q}))))$ 
\STATE $\ell$ = $nn.CrossEntropyLoss()$
\STATE $loss_{FA} = \ell (p_{fa}, p_{pseudo})$
\STATE $loss_{S} = \ell (p_{i,1}, p_{pseudo}) + \ell (p_{i,2}, p_{pseudo})$

\STATE \COMMENT{Final unsupervised loss}
\STATE $\mathcal{L}_u = 0.5*loss_{FA} + 0.5*loss_{S} $ 

\ENDFOR

\end{algorithmic}
\end{algorithm}

In this section, we introduce the basic definitions and the weak-to-strong framework of SSS. Then, we present our Intra-group Consistency Augmentation Framework (ICAF) for the SSS of CdZnTe semiconductors.

\subsection{Problem Definition} \label{Definition}

SSS aims to achieve efficient segmentation by leveraging a vast amount of unlabeled data, thereby minimizing the reliance on extensive pixel-level annotations. A dominant‌ approach for SSS is weak-to-strong consistency regularization~\cite{SSS_Unimatch,SSS_Corrmatch}. This technique applies both weak and strong perturbations to the unlabeled data. Predictions derived from weakly perturbed inputs (e.g., resizing, cropping, flipping) serve as supervisory signals for training on strongly perturbed versions (e.g., color jittering, blurring). Let $\{X_i^u\}^N$ represent $N$ unlabeled images, each with a spatial resolution of $H \times W$. Formally, the traditional consistency regularization can be expressed as:

\begin{align}
    \mathcal{L}_{con} = \frac{1}{|\mathcal{B}|}\!\sum_{i=1}^{|\mathcal{B}|}\!\frac{1}{H\times W}\sum_{j=1}^{H\!\times\!W}\!\ell(p^s_{i(j)}, p^w_{i(j)}) \mathbbm{1}(\max(p^w_{i(j)}) \geq \tau)
    \label{Eq_semi_baseline}
\end{align}
where $\ell$ represents the standard cross-entropy loss, $\mathcal{B}$ denotes the batch size. $p^s_{i(j)}$ and $p^w_{i(j)}$ represent the model's predictions for the strongly and weakly perturbed inputs, respectively. The indicator function $\mathbbm{1}(\cdot)$ ensures that only predictions with confidence scores exceeding the predefined threshold $\tau$ (which defaults to 0.95) are included in the loss calculation, thereby enforcing the training process to focus on high-confidence predictions.

It can be observed that both weak and strong augmentations are derived from the same input image; therefore, applying this paradigm directly in a ``many-to-one'' setting fails to fully exploit the inherent correlations within the data structure.

\subsection{Intra-group Consistency Augmentation Framework (ICAF)} \label{ICAF}

The foundation of effective SSS lies in generating reliable pseudo-labels for unlabeled data while ensuring a balanced learning process between labeled and unlabeled data. Drawing inspiration from Fig.~\ref{image_TPO}, we present a human-inspired solution for CdZnTe semiconductors, utilizing the cross-referencing of multiple images (views) within each group to improve pseudo-label quality, particularly in low-contrast regions. As illustrated in Fig.~\ref{framework}, we revisit the SSS pipeline for CdZnTe semiconductors from a group-oriented perspective and propose ICAF to mine correlations within groups. Specifically, we first establish a novel baseline for SSS of CdZnTe semiconductors by employing Intra-group View Sampling (IVS) to construct consistency constraints within each group. Subsequently, we introduce the Pseudo-label Correction Network (PCN) that comprises the View Augmentation Module (VAM) and the View Correction Module (VCM), leveraging group-wise correlations to enhance consistency representation for each CdZnTe sample.

\noindent
\textbf{Intra-group View Sampling (IVS)} \label{ISS}

The weak-to-strong consistency supervision paradigm encourages the model to produce stable and consistent predictions across various perturbations. Intuitively, we propose a bold \textbf{Hypothesis}: When consistency constraints stem from multiple views within each CdZnTe group, could this ``many-to-one'' data structure naturally serve as a form of data perturbation, enabling the model to learn intrinsic consistent representations across these variations?

To validate this hypothesis, we conducted a straightforward experiment to construct weak and strong perturbations within each group. Mathematical, let $\{X_{i,k}^{u}\}_{1 \leq i \leq N, 1 \leq k \leq 12}$ represent $N$ groups of unlabeled data, where each group contains 12 views. During each training iteration, we randomly sample three views from each group to construct one weak perturbation and two strong perturbations. Formally, Eq. (\ref{Eq_semi_baseline}) is reformulated as:

\begin{equation}
    \mathcal{L}_{con,g} = \frac{1}{|\mathcal{B}|}\!\sum_{i=1}^{|\mathcal{B}|}\!\frac{1}{H\times W}\sum_{j=1}^{H\!\times\!W}\!(\ell_{2,1} + \ell_{3,1}) \mathbbm{1}(\max(p^{w}_{i(j),1}) \geq \tau)
    \label{Eq_group_baseline}
\end{equation}
\begin{equation}
    \ell_{2,1}  = \ell(p^{s}_{i(j),2}, p^{w}_{i(j),1}) 
\end{equation}
\begin{equation}
    \ell_{3,1}  = \ell(p^{s}_{i(j),3}, p^{w}_{i(j),1})
\end{equation}
where $p^{w}_{i(j),1}$ represents the model's predictions for the weakly perturbed inputs derived from the first view, while $p^{s}_{i(j),2}$ and $p^{s}_{i(j),3}$ correspond to the predictions for the strongly perturbed inputs obtained from the second and third views, respectively. 

As demonstrated in Table \ref{table_ablation}, we empirically verify the effectiveness of $\mathcal{L}_{con,g}$ in exploring intrinsic correlations within groups. So far, we have established a group-oriented baseline tailored to the ``many-to-one'' configuration of CdZnTe semiconductors.

\noindent
\textbf{Pseudo-label Correction Network (PCN)} \label{PCN}

\begin{figure}[t]
\centering\includegraphics[scale=0.7]{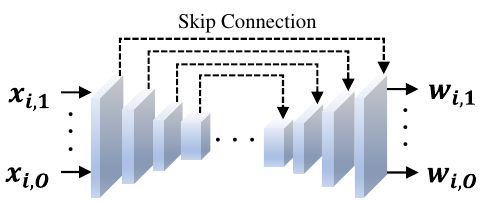}
\caption{Weight Generation Unit (WGU).}
\label{Net1}
\end{figure}

In Eq. (\ref{Eq_group_baseline}), our heuristic group baseline validates the correlations within the CdZnTe group by enforcing consistency constraints. To further strengthen the consistency within each group, we focus on optimizing the quality of pseudo-labels $p^{w}_{i(j),1}$ by explicitly incorporating multiple views. Specifically, we first introduce the View Augmentation Module (VAM), which enhances boundary details by utilizing a weight generator to produce a boundary-aware view $\hat{X}_{i}$. Building upon this enhanced view $\hat{X}_{i}$, we propose the View Correction Module (VCM), which removes redundancy by leveraging an attention mechanism to cross-reference and capture the correlations between $\hat{X}_{i}$ and other views.

\subsubsection{View Augmentation Module (VAM)} \label{VAM}

As shown in Fig. \ref{image_TPO}, images captured under varying illumination angles exhibit complementary boundary information. Evidently, incorporating additional views to capture complete boundary details can enhance the quality of $p^{w}_{i(j),1}$. Various automated data augmentation techniques~\cite{SSS_Augmentation} have significantly improved the performance of SSS. However, these image-level methods typically rely on random perturbations from an augmentation pool applied to individual images, failing to capture the latent relationships among multiple views inherent in the ``many-to-one'' data structure.

Inspired by the effective leveraging of synthetic noisy data to enhance model robustness in pose estimation~\cite{u-generator}, we devise the View Augmentation Module (VAM) to synthesize a boundary-aware view through feature-level aggregation across multiple views. As depicted in Fig. \ref{framework}, we randomly sample $O$ images $X_{i,o}$ from the $i$-th group and apply image transformations $T_w(\cdot)$, such as resizing, cropping, and flipping to obtain $x_{i,o}$. These transformed views $x_{i,o}$ are then fed into the Weight Generation Unit (WGU) to dynamically guide the data augmentation process within each group. As shown in Fig.~\ref{Net1}, the WGU is an encoder-decoder-based unit that adaptively integrates shallow and deep features to dynamically assign weights to input views, generating an additional boundary-aware view $\hat{X}_{i}$ online during each iteration. Formally:

\begin{equation}
    \hat{X}_{i} = \sum_{o=1}^{O} w_{i,o} \cdot x_{i,o}, \quad \text{where } w_{i,o} = WGU(x_{i,o}).
\end{equation}

\begin{equation}
    WGU(x_{i,o}) = \text{Softmax} (f_{de} (f_{en}(x_{i,o}; \theta_{en}) ; \theta_{de} ))
\end{equation}
where $w_{i,o}$ is the weight of $x_{i,o}$, $f_{en}$ and $f_{de}$ are the encoder and decoder functions of WGU, parameterized by $\theta_{en}$ and $\theta_{de}$.

Therefore, we aggregate boundary details from $O$ images within each group to form $\hat{X}_{i}$, thereby improving the quality of pseudo-labels for further training. 

\begin{figure}[t]
\centering\includegraphics[scale=0.7]{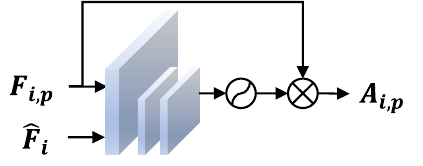}
\caption{Spatial Interaction Unit (SIU).}
\label{Net2}
\end{figure}

\subsubsection{View Correction Module (VCM)} \label{VCM}

Considering that $\hat{X}_{i}$ contains rich boundary clues crucial for view correction, we employ the Spatial Interaction Unit (SIU) to harness these discriminative features by forming pairs between $\hat{X}_{i}$ and other views for information interaction. Motivated by the spatial attention mechanism~\cite{CBAM}, SIU dynamically adjusts responses at each feature map position. Specifically, SIU randomly selects $P$ views $X_{i,p}$ from the set of $O$, maps both $X_{i,p}$ and $\hat{X}_{i}$ to the embedding space:

\begin{equation} 
F_{i,p} = En(T_w(X_{i,p})), \quad \text{for } p = 1, \dots, P 
\end{equation}

\begin{equation}
\hat{F}_{i} = En(\hat{X}_{i})
\end{equation} 
where $En(\cdot)$ represents the encoder to derive the latent feature mappings.

As depicted in Fig.~\ref{Net2}, we pair each view $F_{i,p}$ with $\hat{F}_{i}$ and capture the relationship in each pair to highlight salient regions and filter out irrelevant information from $F_{i,p}$. Formally:

\begin{equation} 
A_{i,p} = \sigma(Proj(Cat(F_{i,p}, \hat{F}_{i}))) \times F_{i,p} 
\end{equation}
where $Cat(\cdot)$ denotes the concatenation operation, $Proj(\cdot)$ serves as a channel integrator implemented with convolutional layers, $\sigma$ represents the Sigmoid function. Then, a summation is performed on $A_{i,p}$ for $p = 1, \dots, P$ to obtain the corrected clues: $A_i = \sum_{p=1}^{P} A_{i,p}$. 

So far, $A_{i}$ has been derived by cross-referencing $\hat{X}_{i}$ with $P$ views from each group. Furthermore, we sample $Q$ views from the set $O$ to supplement the strongly perturbed inputs in Eq. (\ref{Eq_group_baseline}). This constitutes the ICAF for the multi-input CdZnTe consistency regularization. 

In conclusion, the strength of the ICAF approach lies in its ability to extract meaningful clues within groups. The noval baseline establishes a consistency framework from a group perspective, and the PCN further enhances the accuracy of the supervision signal within this paradigm by enhancing boundary awareness and facilitating view interaction. This allows for the progressive refinement of pseudo-labels and helps mitigate confirmation bias.

\subsection{Loss Function} \label{Loss Function}

Overall, we present the PCN that incorporates VAM and VCM to enhance the quality of pseudo-labels, which are used to supervise $Q$ strongly perturbed views. Furthermore, we also integrated a dropout layer for additional feature perturbation. Eq. (\ref{Eq_group_baseline}) is reformulated as $\mathcal{L}_u$:

\begin{equation}
    \mathcal{L}_u = \frac{1}{|\mathcal{B}|}\!\sum_{i=1}^{|\mathcal{B}|}(0.5\times\ell_{S} + 0.5\times\ell_{FA}) \mathbbm{1}(\max(PCN(p^{w}_{i(j),O})) \geq \tau)
\end{equation}

\begin{equation}
    \ell_{S}  = \frac{1}{Q}(\ell(p^{s}_{i(j),2}, PCN(p^{w}_{i(j),O})+ ... + \ell(p^{s}_{i(j),Q}, PCN(p^{w}_{i(j),O}))
\end{equation}

\begin{equation}
    \ell_{FA}  = \ell(dropout(PCN(p^{w}_{i(j),O})), PCN(p^{w}_{i(j),O}))
\end{equation}
where 0.5 is the default coefficient, $\ell_{FA}$ denotes the feature augmentation loss for enforcing consistency restrictions, and $\ell_{S}$ represents the strongly perturbed views.

For labeled data, we also utilize the PCN to perform cross-reference augmentation during each iteration, ensuring a balanced learning process between labeled and unlabeled data. Similar to $\{X_{i,k}^{u}\}_{1 \leq i \leq N, 1 \leq k \leq 12}$, let $\{X_{i,k}^{l}, Y_{i}^l\}_{1 \leq i \leq M, 1 \leq k \leq 12}$ denote $M$ groups of labeled data, where each group contains 12 views and $Y_i^l$ is the corresponding GT segmentation map, each also with a spatial resolution of $H \times W$. Here, $Y_i^l \in \mathbb{R}^{C \times H \times W}$ is a one-hot encoded label map, where $C$ represents the number of segmentation classes, and $N \gg M$ in practical industrial contexts. The supervised loss $\mathcal{L}_l$ is defined as:

\begin{equation}
    \mathcal{L}_l = \frac{1}{|\mathcal{B}|} \sum_{i=1}^{|\mathcal{B}|} \ell(PCN(X_{i(j),O}^{l}), Y_{i(j)}^{l})
\end{equation}
where $X_{i(j),O}^{l}$ is the sampled $O$ labeled views.

The objective of SSS is to jointly optimize the supervised loss $\mathcal{L}_l$ and the unsupervised loss $\mathcal{L}_u$. The total training loss is formulated as:

\begin{equation} 
\mathcal{L}_{\text{total}} = \mathcal{L}_l + \lambda \mathcal{L}_u, 
\end{equation} 
where $\lambda$ is a trade-off coefficient and is set to 0.5 following~\cite{SSS_Unimatch,SSS_Corrmatch}. 

It is crucial to emphasize that the entire ICAF framework is employed exclusively during the training phase. The pseudocode for processing unlabeled data during training is provided in Algorithm~\ref{algorithm}. During the testing phase, the ``many-to-one'' structure is transformed into the standard ``one-to-one'' configuration for evaluation.

\section{EXPERIMENTAL RESULTS AND DISCUSSIONS}
In this section, we evaluate the performance of the proposed ICAF approach on the TPO dataset. All experiments are conducted using PyTorch and implemented on a system equipped with four NVIDIA 4090 GPUs.

\subsection{Dataset Overview} \label{sec_TPO}

\begin{figure}[t]
\centering
    \includegraphics[scale=0.6]{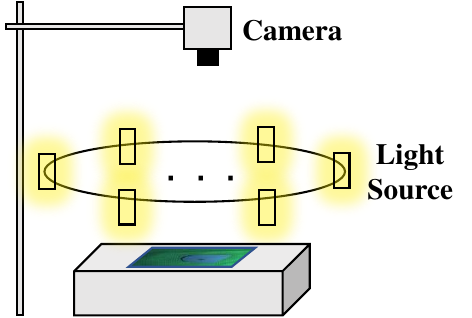}
\caption{Image acquisition setup. Leveraging a single-camera, multi-light source system, we capture a group of RGB images of CdZnTe under 360-degree illumination conditions.
\label{image_acqu}}
\end{figure}

The experiments were conducted using the self-collected Twelve images Plus One corresponding Label (TPO) dataset, gathered through the image acquisition setup depicted in Fig. \ref{image_acqu}. Each group contains 12 RGB images of a CdZnTe semiconductor captured under varying lighting angles, all sharing a common GT label. To balance training costs with model generalization capabilities, the dataset was augmented, resulting in 410 groups of images for training and 894 groups for testing. Further details on the TPO dataset can be found in~\cite{PCKA}.

\subsection{Implementation Details}

To ensure fair comparisons with previous state-of-the-art SSS methods, we adopt DeepLabV3+~\cite{v3plus} with the ResNet-101~\cite{resnet} backbone pre-trained on ImageNet~\cite{imagenet}. Following~\cite{SSS_Unimatch}, the ResNet backbone is configured with an output stride of 16 for all experiments. Each training mini-batch consists of 16 groups, comprising 8 labeled and 8 unlabeled groups. The initial learning rate is set to 0.001, using a SGD optimizer with momentum, and the model is trained for 80 epochs with a poly decay learning rate scheduler. The image resolution is fixed at 320×320 pixels.

For evaluation, we report the mean Intersection-over-Union (mIoU), a widely used metric in semantic segmentation that quantifies the overlap between predicted and GT segmentations. All results are computed on the TPO test set with single-scale inference, ensuring consistency and comparability with previous studies~\cite{SSS_Unimatch, SSS_Corrmatch}.
\subsection{Comparison With State-of-the-Art Methods}

We conduct comparative experiments with state-of-the-art SSS algorithms, UniMatch~\cite{SSS_Unimatch} and CorrMatch~\cite{SSS_Corrmatch}, to validate the effectiveness of our ICAF approach for CdZnTe semiconductors. We adhere strictly to the default configurations provide in the official code. The results for these algorithms are obtained by retraining their official implementations on the TPO dataset. Our experiments use standard dataset partitions of 5\%, 10\%, 25\%, and 50\%. Notably, to thoroughly validate the effectiveness of our group-based approach, we also include an extreme partition with a labeled ratio of 5\textperthousand.

As shown in Table \ref{table_main}, our ICAF consistently outperforms the compared methods. Compared to the supervised baseline, our approach achieves improvements of +8.7\%, +7.1\%, +5\%, +4.4\%, and +3.1\% across the different data partitions. The performance advantage is particularly pronounced when using fewer labeled data for training, achieving outstanding results even with just 5\textperthousand~of labeled data. Additionally, our method surpasses the best previous results by 1.9\%, 3.7\%, 3.4\%, 1.8\%, and 1.7\% on each dataset split. These results clearly highlight the remarkable performance of our group-oriented solution.

\begin{table}[t]
\renewcommand{\arraystretch}{1}
\caption{Performance comparison on the TPO test set. ``Bg'', ``Cd'', and ``De'' refer to the background, monocrystalline CdZnTe, and defects, respectively. The numbers 2, 20, 40, 103, and 205 indicate the number of CdZnTe image groups used in the experiments.}
\label{table_main}
\centering 
\resizebox{\linewidth}{!}{						
\begin{tabular}{ l  c  c  c  c  c}			
	\hline \\[-3mm]
    \multirow{2}{*}{Partition}  & \multirow{2}{*}{Method}                       &\multirow{2}{*}{mIoU(\%)}  & \multicolumn{3}{c}{IoU}                                        \\ \cline{4-6}
                                &                                               &                           & Bg                       & Cd              & De                \\ \hline
                                & Supervised                                    & 61.9                      & 99.1                     & 72.5            & 14.2              \\ 
    5\textperthousand           & UniMatch~\cite{SSS_Unimatch}~\pub{CVPR'23}    & 68.7                      & \textbf{99.6}            & 71.7            & 34.9              \\ 
    (2)                         & CorrMatch~\cite{SSS_Corrmatch}~\pub{CVPR'24}  & 66.4                      & 98.0                     & 73.5            & 27.6              \\ 
                                & Ours                                          & \textbf{70.6}             & \textbf{99.6}            & \textbf{76.3}   & \textbf{35.9}     \\ \hline
                                & Supervised                                    & 72.7                      & \textbf{99.7}            & 79.3            & 39.2              \\ 
    5\%                         & UniMatch~\cite{SSS_Unimatch}~\pub{CVPR'23}    & 76.1                      & 99.1                     & 80.4            & 48.7              \\ 
    (20)                        & CorrMatch~\cite{SSS_Corrmatch}~\pub{CVPR'24}  & 74.5                      & \textbf{99.7}            & 80.1            & 43.6              \\ 
                                & Ours                                          & \textbf{79.8}             & \textbf{99.7}            & \textbf{86.2}   & \textbf{53.6}     \\ \hline
                                & Supervised                                    & 77.0                      & \textbf{99.8}            & 84.3            & 46.8              \\ 
    10\%                        & UniMatch~\cite{SSS_Unimatch}~\pub{CVPR'23}    & 77.3                      & 99.7                     & 82.9            & 49.4              \\ 
    (40)                        & CorrMatch~\cite{SSS_Corrmatch}~\pub{CVPR'24}  & 78.6                      & 99.5                     & 84.6            & 51.7              \\ 
                                & Ours                                          & \textbf{82.0}             & \textbf{99.8}            & \textbf{87.4}   & \textbf{58.8}     \\ \hline
                                & Supervised                                    & 80.1                      & \textbf{99.8}            & 85.8            & 54.6              \\ 
    25\%                        & UniMatch~\cite{SSS_Unimatch}~\pub{CVPR'23}    & 82.3                      & 99.6                     & 87.7            & 59.6              \\ 
    (103)                       & CorrMatch~\cite{SSS_Corrmatch}~\pub{CVPR'24}  & 82.7                      & \textbf{99.8}            & 88.6            & 59.6              \\ 
                                & Ours                                          & \textbf{84.5}             & \textbf{99.8}            & \textbf{90.0}   & \textbf{63.6}     \\ \hline
                                & Supervised                                    & 83.2                      & \textbf{99.8}            & 88.8            & 61.0              \\ 
    50\%                        & UniMatch~\cite{SSS_Unimatch}~\pub{CVPR'23}    & 84.4                      & 99.8                     & 89.9            & 63.4              \\ 
    (205)                       & CorrMatch~\cite{SSS_Corrmatch}~\pub{CVPR'24}  & 84.6                      & 99.8                     & 89.6            & 64.3              \\ 
                                & Ours                                          & \textbf{86.3}             & \textbf{99.8}            & \textbf{91.0}   & \textbf{68.1}     \\ \hline

\end{tabular}}			
\end{table}

\newcommand{\methodName}{\textit\textbf{{$PCN$}}}{}
\begin{table}[t]
\caption{Analysis of each component in ICAF.}
\centering
\resizebox{\linewidth}{!}{
\begin{tabular}{ccc|cc|cc|c}

 & \multicolumn{2}{c|}{\textit\textbf{{Baseline}}}   & \multicolumn{2}{c|}{\textit\textbf{{Augmentation}}}    & \multicolumn{2}{c|}{\methodName{}}    & \multirow{2}{*}{mIoU(\%)}    \\ 
 &    Semi                        &     Group                         & CA             & FA                                   & VAM          & VCM                    &                              \\ \midrule
1& \checkmark                     &                                   &                &                                      &              &                        & 81.1                         \\ \midrule
2&                                & \checkmark                        &                &                                      &              &                        & 82.5                         \\
3&                                & \checkmark                        & \checkmark     &                                      &              &                        & 83.3                         \\
4&                                & \checkmark                        & \checkmark     & \checkmark                           &              &                        & 83.9                         \\
5&                                & \checkmark                        & \checkmark     & \checkmark                           & \checkmark   &                        & 85.2                         \\ 
6&                                & \checkmark                        & \checkmark     & \checkmark                           & \checkmark   & \checkmark             & \textbf{86.3}                \\ 
    \bottomrule                     
    \end{tabular}}
    \label{table_ablation}
\end{table}

\begin{table}[t]
	\caption{The segmentation results obtained using different values of $O$ and $P$ to refine the pseudo-labels.}
	\centering
	\begin{tabular}{c|cccccc}
    \toprule
    \diagbox{$O$}{$P$}           & 1            & 3                & 6       & 9      \\
    \midrule                        
                          1      & 81.9         & 82.9             & 84.9    & 81.0   \\ 
                          3      & 83.9         & 85.6             & 84.7    & 81.2   \\
                          6      & 84.6         & \textbf{86.3}    & 84.5    & 82.2   \\
                          9      & 83.4         & 85.7             & 83.6    & 83.3   \\
    \bottomrule
    \end{tabular}
    \label{table_ablation_OP}
\end{table}

\begin{table}[t]
	\caption{The segmentation results obtained using different values of $Q$ for strongly perturbed reference images.}
	\centering
	\begin{tabular}{ccc|c}
		\toprule
        
		           O      & P     & Q    & mIoU(\%)       \\
		
		\midrule
                      6       & 3    & 1    & 82.9           \\ 
                      6       & 3    & 2    & \textbf{86.3}  \\
		           6       & 3    & 3    & 85.5           \\ 
		\bottomrule
	\end{tabular}

     \label{table_ablation_Q}
\end{table}

\subsection{Ablation Studies} 

In this subsection, we evaluate the contribution of the proposed components and the parameter effectiveness under the 50\% dataset partition protocol.

\subsubsection{Components effectiveness of ICAF}

We provide a comprehensive evaluation of the contribution of each component within the ICAF framework. The first row of Table \ref{table_ablation} shows the results of the conventional weak-to-strong baseline, as described in Eq. (\ref{Eq_semi_baseline}), referred to as the Semi-Baseline. The second row presents the newly proposed baseline, detailed in Eq. (\ref{Eq_group_baseline}), termed the Group-Baseline. As shown in Table \ref{table_ablation}, all proposed components significantly contribute to enhance the overall performance of ICAF. Initially, the mIoU increases from 81.1\% to 82.5\% with the adoption of the Group-Baseline, validating the effectiveness of our proposed intra-group consistency hypothesis. Subsequently, we employ commonly used data augmentation techniques in SSS: Cutmix-based image augmentation (denoted as CA) and feature augmentation using Dropout (denoted as FA). The results show performance improvements to 83.3\% and 83.9\%, respectively, demonstrating that the data augmentation methods enhance the inherent consistency learning process within a group. The introduction of our proposed VAM for optimizing pseudo-labels then leads to a substantial performance increase to 85.2\%. Finally, our complete configuration further improves performance to 86.3\%, validating the overall effectiveness of our approach.

In summary, the new baseline introduces consistency constraints to establish a robust foundation for pseudo-label correction from a group perspective; the VAM module aggregates boundary information by generating augmented views; and the VCM module further removes noise through inter-view interactions.

\subsubsection{Hyper-parameter ablation}

The hyper-parameters $O$, $P$, and $Q$ play crucial roles in determining the quality of pseudo-labels and controlling the predefined perturbation space. 

As shown in Table \ref{table_ablation_OP}, we identify the optimal values for $O$ and $P$, which correspond to the ideal number of views for boundary enhancement in VAM and the number of view pairs for information interaction in VCM, respectively. First, we analyze the sampling quantities of $O$ and $P$ separately: Whether for $O$ or $P$, as the sampling quantity increases, the network captures richer boundary information from more views, leading to corresponding performance gains (81.9\% to 84.9\%). However, once the number of sampled views exceeds a certain threshold, the model's performance begins to decline (84.9\% to 81.0\%), likely due to the additional noise introduced by more blurred boundaries. Next, we examine the combined state of $O$ and $P$, where the optimal parameter configuration appears in the lower triangular region of the table (mIoU = 86.3\%), suggesting that $O$ should be greater than $P$. An intuitive explanation is that the boundary-aware view should contain more diverse and discriminative features than the cross-referenced views in order to effectively extract useful clues. After determining the optimal values for $O$ and $P$ as 6 and 3, respectively, we further conducted experiments to determine the optimal perturbation space $Q$, as presented in Table \ref{table_ablation_Q}. Similar to the previous experimental results, the value of $Q$ reaches saturation when set to 2. Based on our findings, we selected $O = 6$, $P = 3$, and $Q = 2$ as the optimal configuration for our approach.

\subsection{Qualitative Studies}
\subsubsection{Visualization examples for ICAF}

\begin{figure}[t]
\centering
    \includegraphics[scale=0.6]{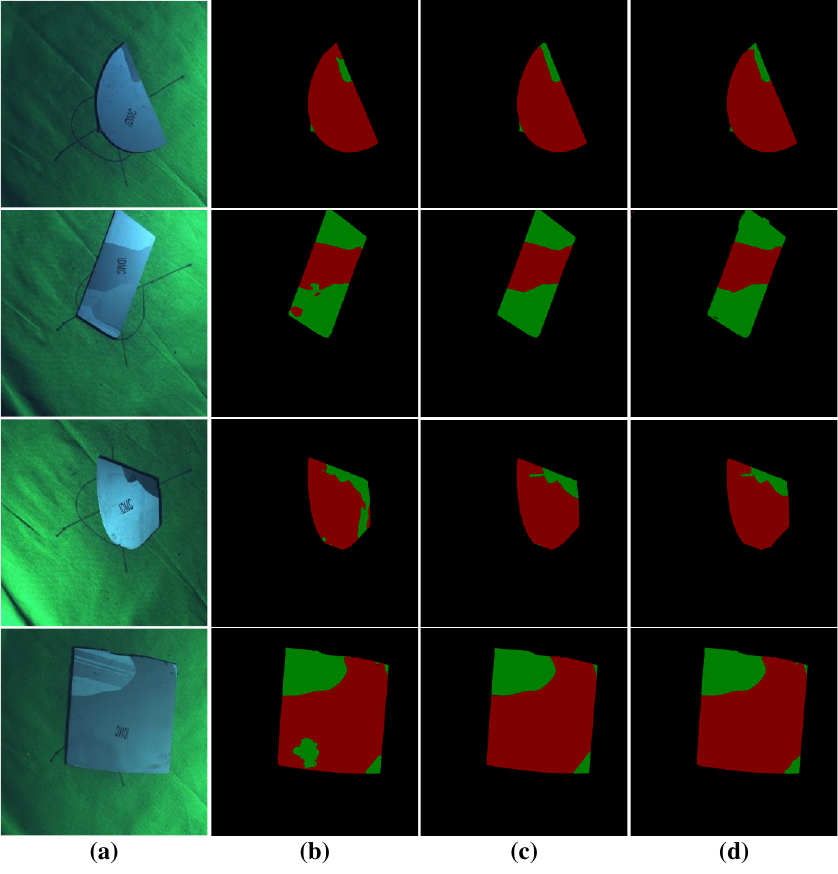}
\caption{Qualitative results from the test set of the TPO dataset: (a) Input RGB image; (b) Predictions from the supervised-only method; (c) GT; (d) Predictions from the proposed ICAF approach.}
\label{fig_main}
\end{figure}

In Fig. \ref{fig_main}, we present a visual comparison between the proposed ICAF algorithm and the supervised-only baseline. It is evident that our solution is more visually appealing compared to the baseline method. Furthermore, the baseline method tends to misclassify connected regions as evidenced by the qualitative comparison in the last row, whereas our approach alleviates this issue by employing pseudo-label correction strategies that leverage multiple input views.

\begin{figure}[t]
\centering
    \includegraphics[scale=0.6]{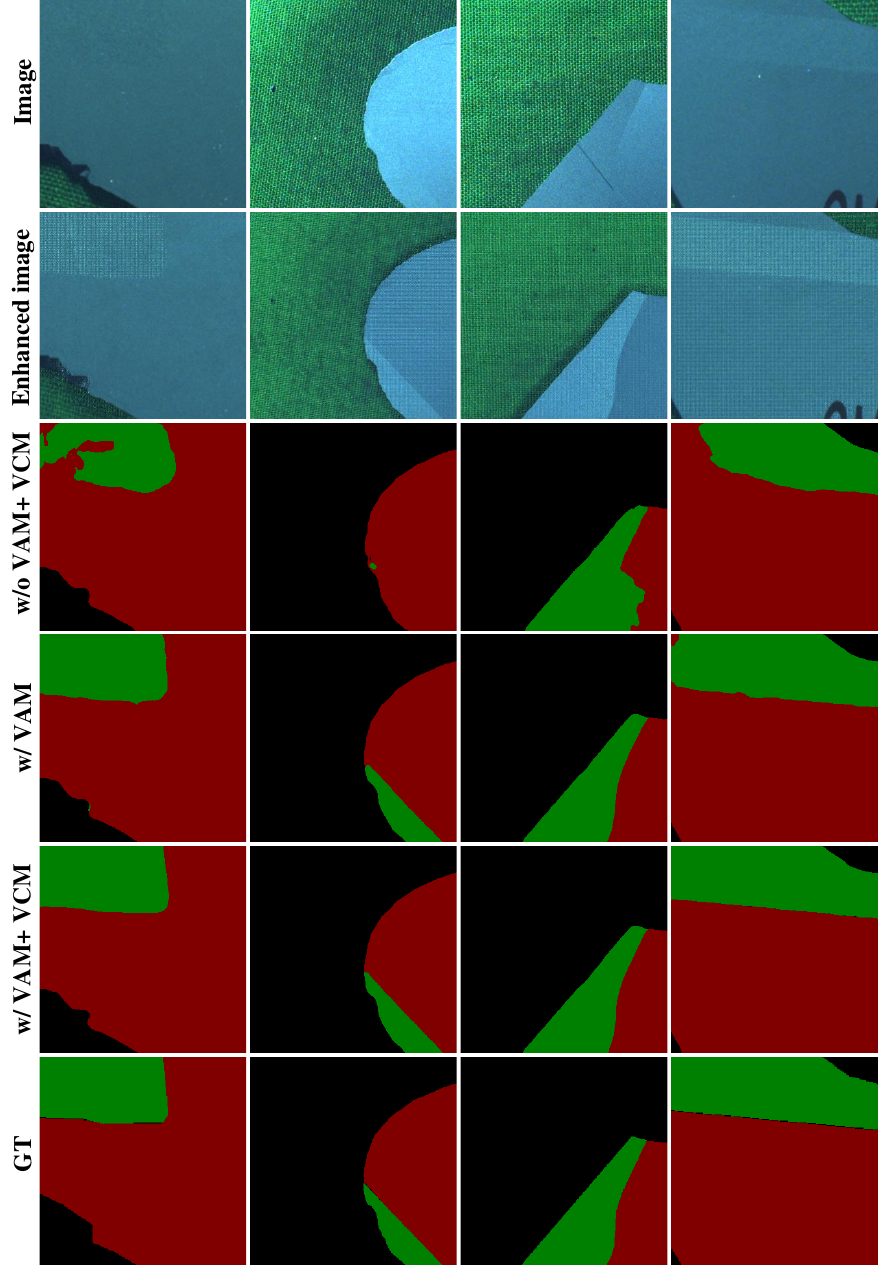}
\caption{Visualization of the pseudo-label correction process.}
\label{fig_pseudo}
\end{figure}

\subsubsection{The visualization of the pseudo-label correction}

To provide a more intuitive illustration of the effectiveness of our pseudo-label correction strategies during training, we present visualization results corresponding to Table \ref{table_ablation}. 

In the low-contrast regions of the CdZnTe images, both the model and the human eye struggle to recognize defect boundaries from a single view, as depicted in the first row of Fig. \ref{fig_pseudo}. Consequently, ensuring the quality of pseudo-labels becomes challenging, as shown in the third row. With the introduction of VAM, it becomes evident that the VAM effectively highlights boundary details in low-contrast regions by integrating information from multiple views, significantly refining previously blurred boundaries and correcting misclassified categories in the pseudo-labels, as shown in the second and fourth rows. Finally, by examining the last two rows, the pseudo-label undergoes further refinement through the interaction of information between the boundary-aware view and other views, resulting in a pseudo-label that closely approximates the GT. It is clear that our pseudo-label correction framework substantially improves both the quantity and connectivity of pixels classified as pseudo-labels, highlighting the effectiveness of our method in mitigating the impact of confirmation bias.

\section{Conclusion}

Low-contrast regions in industrial defect imaging pose a challenge for large-scale labeled datasets in automated defect detection. Traditional SSS methods are unable to address the ``many-to-one'' characteristics of the CdZnTe semiconductors, resulting in suboptimal performance. To address this issue, we revisit the SSS pipeline from a group-oriented perspective and propose an innovative, human-inspired multi-input consistency regularization approach for CdZnTe, called ICAF. Specifically, IVS establishes consistency constraints based on the group-oriented baseline, and the PCN incorporates VAM and VCM further enhances the consistency characterization. VAM generate a boundary-aware view to refine the boundaries, and 
VCM focuses on the interaction of information between the newly generated view and other views to eliminate redundancy. This method effectively alleviates annotation difficulties arising from variations in illumination, offering a practical solution for industrial defect labeling. We hope this approach will provide valuable technical insights and robust support for real-world industrial inspection applications.
\ifCLASSOPTIONcaptionsoff
\newpage
\fi

\bibliographystyle{IEEEtran}
\bibliography{IEEEabrv,lph}

\newpage

\begin{IEEEbiography}[{\includegraphics[width=1in,height=1.25in,clip,keepaspectratio]{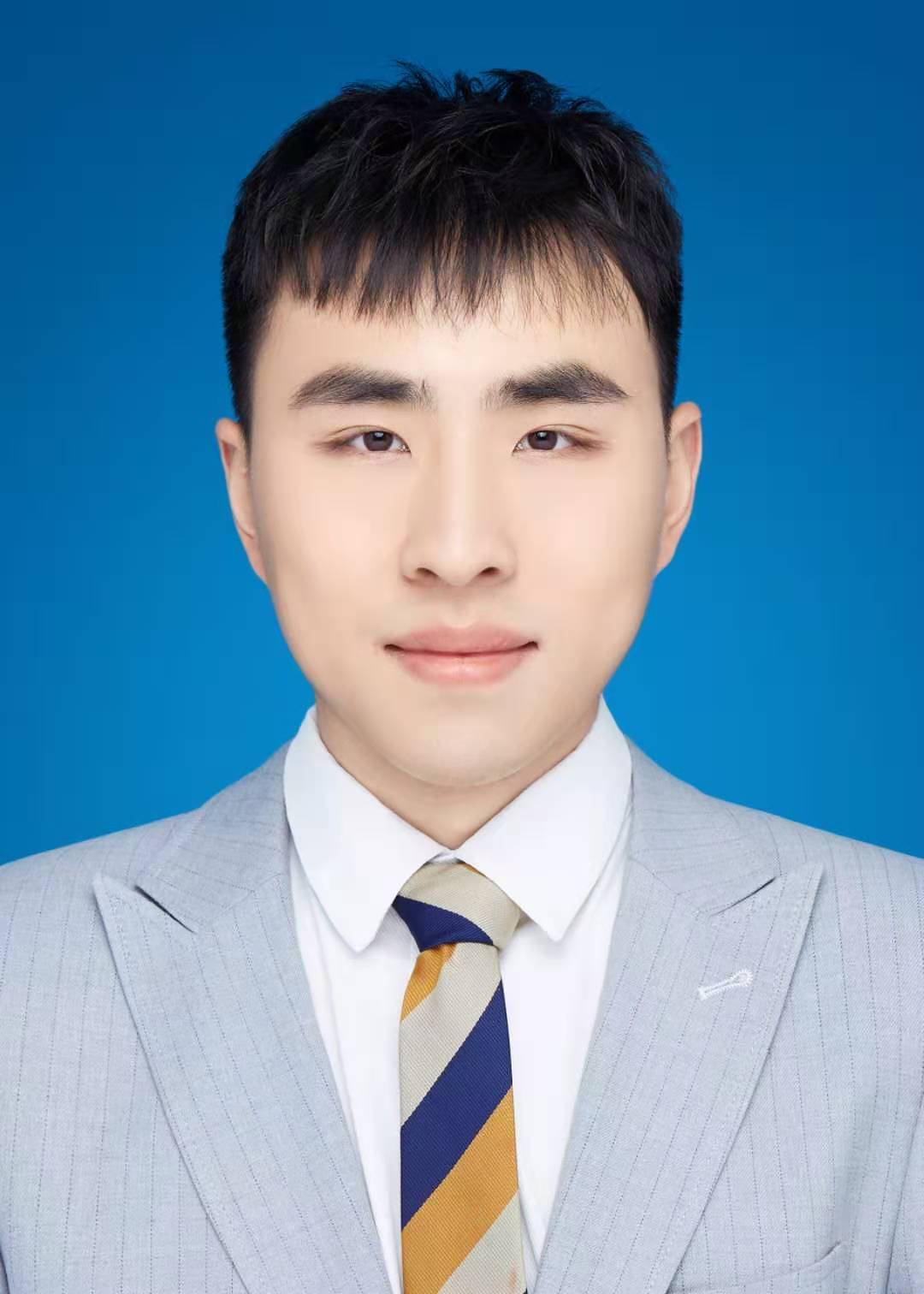}}]
{Peihao Li} is currently pursuing the Ph.D. degree in information and communication engineering with the Institute of Information Science, Beijing Jiaotong University, Beijing, China. His research interests include semantic segmentation and computer vision.
\end{IEEEbiography}

\begin{IEEEbiography}[{\includegraphics[width=1in,height=1.25in,clip,keepaspectratio]{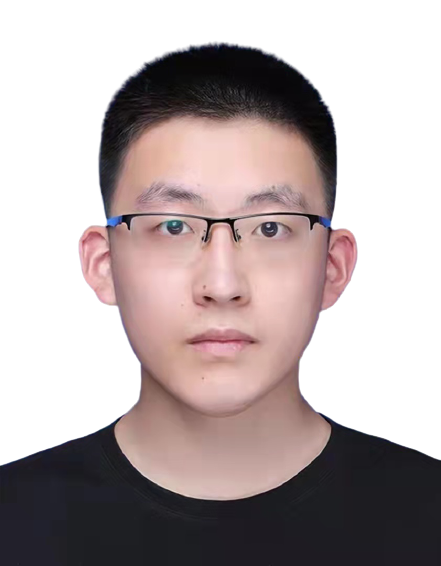}}]
{Yan Fang} is currently a third-year PhD student in Information and Communication Engineering at the Institute of Information Science, Beijing Jiaotong University, Beijing, China. His research interests encompass a broad range of topics within computer vision, efficient learning, and their intersections, including semi-supervised learning, continual learning, and the development of efficient vision foundation models.
\end{IEEEbiography}

\begin{IEEEbiography}
[{\includegraphics[width=1in,height=1.25in,clip,keepaspectratio]{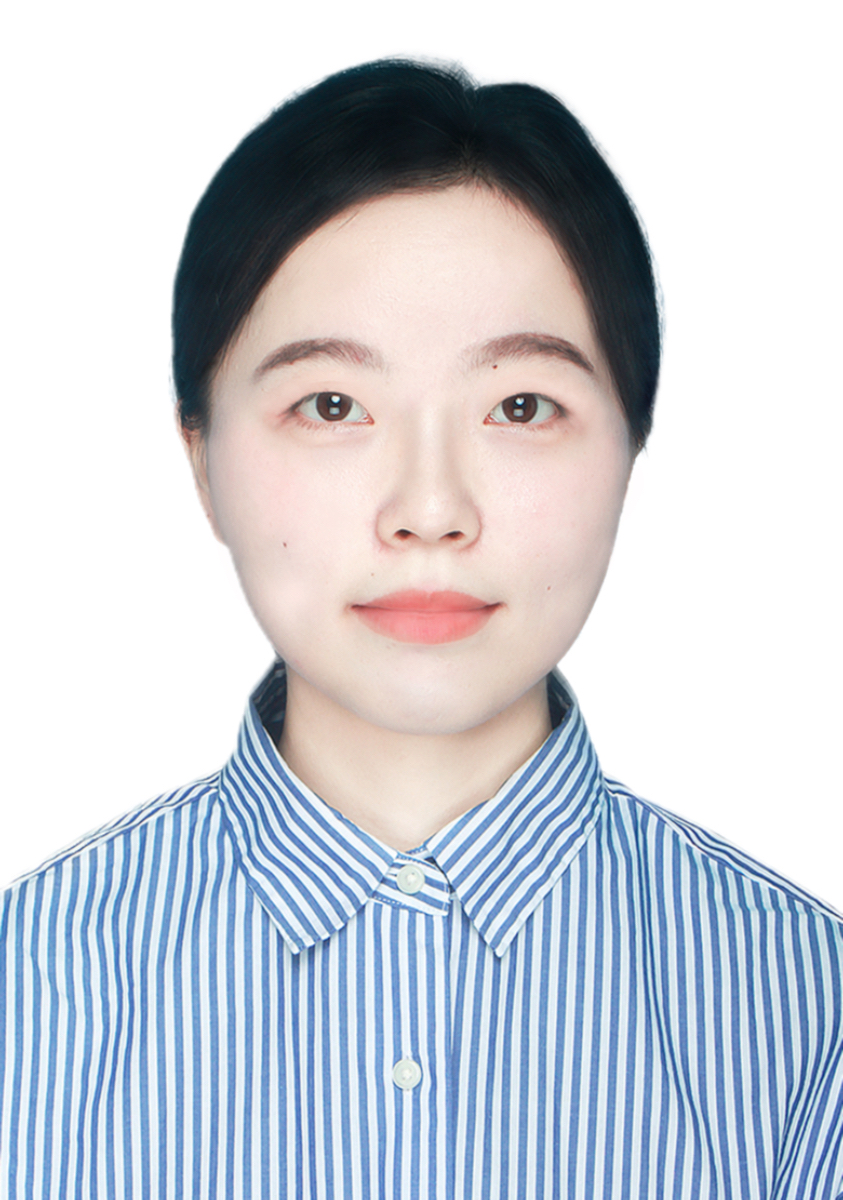}}]
{Man Liu} 
received her M.S. degree from Anhui University of Technology, China, in 2020 and her Ph.D. degree from Beijing Jiaotong University, China, in 2025. She is currently a Lecturer with the School of Artificial Intelligence, Anhui University, Hefei, China. Her current research interests include image classification, zero-shot learning, visual-and-language learning.
\end{IEEEbiography}

\begin{IEEEbiography}
[{\includegraphics[width=1in,height=1.25in,clip,keepaspectratio]{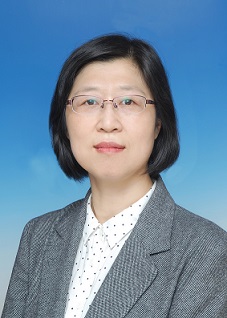}}]
{Huihui Bai}
received her B.S. degree from Beijing Jiaotong University, China, in 2001, and her Ph.D. degree from Beijing Jiaotong University, China, in 2008. She is currently a professor in Beijing Jiaotong University. She has been engaged in R \& D work in video coding technologies and standards, such as HEVC, 3D video compression, multiple description video coding (MDC), and distributed video coding (DVC).
\end{IEEEbiography}

\begin{IEEEbiography}[{\includegraphics[width=1in,height=1.25in,clip,keepaspectratio]{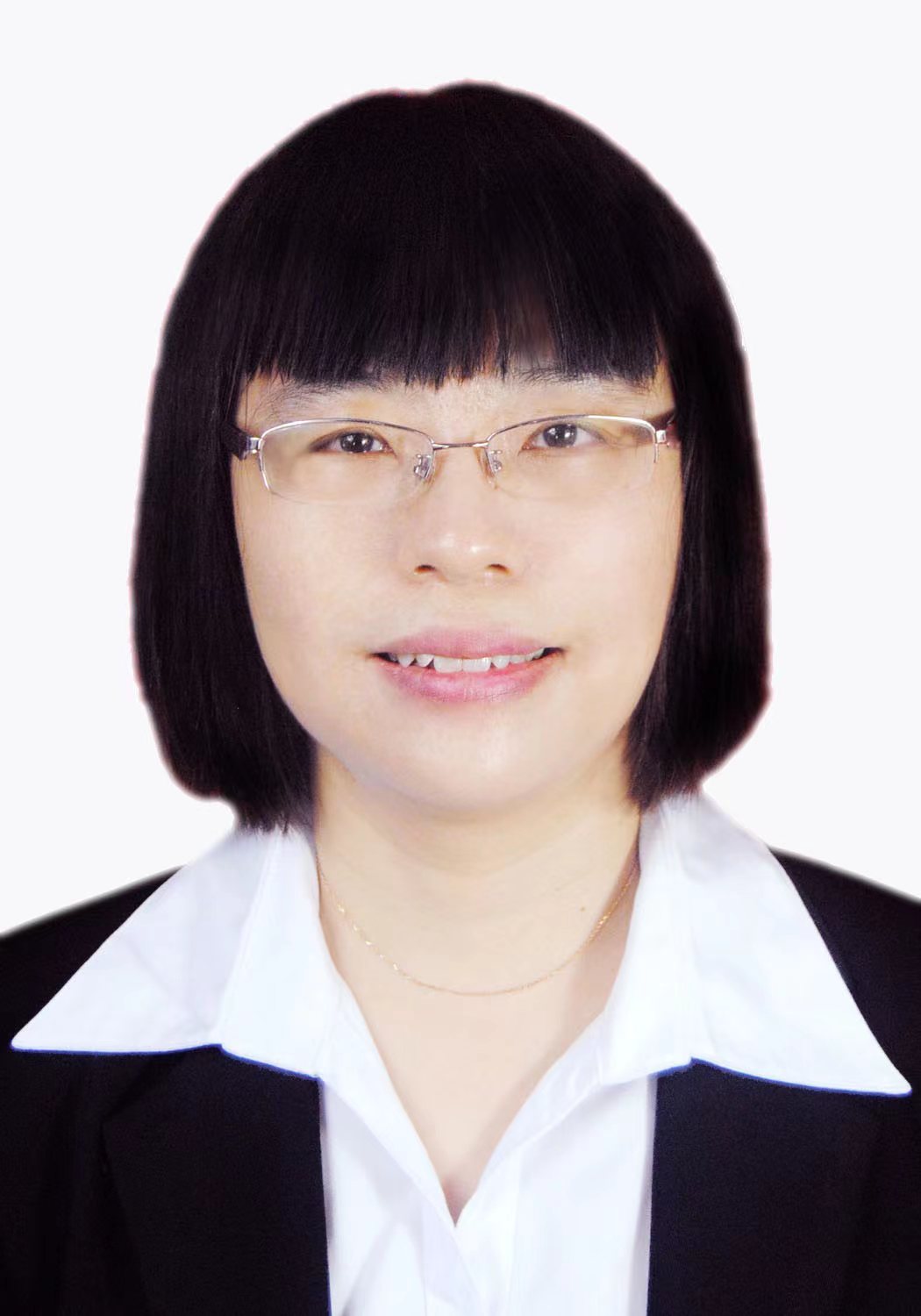}}]
{AnHong Wang} received the Ph.D. degree from the Institute of Information Science, Beijing Jiaotong University (BJTU), in 2009. She became an Associate Professor with TYUST, in 2005, where she became a Professor, in 2009. She is currently the Director of the Institute of Digital Media and Communication,
Taiyuan University of Science and Technology. Her research interests include image and video coding and secret image sharing.
\end{IEEEbiography}

\begin{IEEEbiography}[{\includegraphics[width=1in,height=1.25in,clip,keepaspectratio]{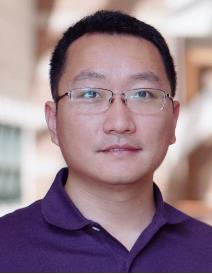}}]
{Yunchao Wei} received the PhD degree from Beijing Jiaotong University, Beijing, China, in 2016. He is currently a professor with the Center of Digital Media Information Processing, Institute of Information Science, Beijing Jiaotong University. He was a Postdoctoral Researcher at Beckman Institute, UIUC, from 2017 to 2019. He is ARC Discovery Early Career Researcher Award Fellow from 2019 to 2021. His current research interests include computer vision and machine learning.
\end{IEEEbiography}

\begin{IEEEbiography}
[{\includegraphics[width=1in,height=1.25in,clip,keepaspectratio]{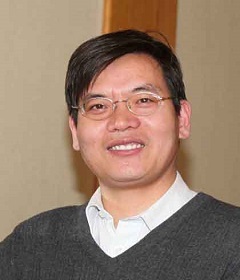}}]
{Yao Zhao}
(Fellow, IEEE) received the BS degree from Fuzhou University, China, in 1989, and the ME degree from Southeast University, Nanjing, China, in 1992, both from the Radio Engineering Department, and the PhD degree from the Institute of Information Science, Beijing Jiaotong University (BJTU), China, in 1996. He is currently the director of the Institute of Information Science, BJTU. His current research interests include image/video coding, digital watermarking and forensics, and video analysis and understanding. He serves on the editorial boards of several international journals, including as associate editors of IEEE Transactions on Cybernetics, IEEE Signal Processing Letters, and an area editor of Signal Processing: Image Communication (Elsevier), etc. He was named a distinguished young scholar by the National Science Foundation of China in 2010 and was elected as a Chang Jiang Scholar of Ministry of Education of China in 2013. 
\end{IEEEbiography}

\end{document}